\documentclass[letterpaper, 10 pt, conference]{ieeeconf}   
\IEEEoverridecommandlockouts                             
\title{\Large \bf Learning Concept-Based Causal Transition and Symbolic Reasoning \\ for Visual Planning}

\let\labelindent\relax
\usepackage{graphicx} \graphicspath{ {figures/} }
\usepackage{amsmath,amssymb,mathabx}
\usepackage[utf8]{inputenc} 
\usepackage[T1]{fontenc}    
\usepackage{url}
\usepackage{algorithmic}
\usepackage[linesnumbered,ruled,vlined]{algorithm2e}
\usepackage{times}
\usepackage{acronym}
\usepackage{balance}
\usepackage{xspace}
\usepackage{setspace}
\usepackage[skip=3pt,font=small]{subcaption}
\usepackage[skip=3pt,font=small]{caption}
\usepackage[dvipsnames,svgnames,x11names]{xcolor}
\usepackage[capitalise,nameinlink]{cleveref}
\usepackage{booktabs,tabularx,colortbl,multirow,multicol,array,makecell}
\usepackage{cite}
\usepackage{bbm,bm}
\usepackage{epigraph}
\usepackage{colortbl}
\usepackage[misc]{ifsym}

\makeatletter
\DeclareRobustCommand\onedot{\futurelet\@let@token\@onedot}
\def\@onedot{\ifx\@let@token.\else.\null\fi\xspace}
\def\eg{\emph{e.g}\onedot} 

\def\ie{\emph{i.e}\onedot}

\makeatother

\makeatletter
\def\BState{\State\hskip-\ALG@thistlm}
\makeatother

\makeatletter
\renewcommand{\paragraph}{%
  \@startsection{paragraph}{4}%
  {\z@}{0ex \@plus 0ex \@minus 0ex}{-1em}%
  {\hskip\parindent\normalfont\normalsize\bfseries}%
}
\makeatother

\definecolor{gblue}{HTML}{4285F4}
\definecolor{gred}{HTML}{DB4437}
\definecolor{ggreen}{HTML}{0F9D58}

\definecolor{mygray}{gray}{.92}

\definecolor{mygray}{gray}{.92}

\newcolumntype{a}{>{\columncolor{mygray}}c}
\newcommand{\thickhline}{%
    \noalign {\ifnum 0=`}\fi \hrule height 1pt
    \futurelet \reserved@a \@xhline
}

\newcommand \footnoteONLYtext[1]
{
	\let \mybackup \thefootnote
	\let \thefootnote \relax
	\footnotetext{#1}
	\let \thefootnote \mybackup
	\let \mybackup \imareallyundefinedcommand
}

\crefname{section}{Sec.}{Secs.}
\Crefname{section}{Section}{Sections}
\Crefname{table}{Table}{Tables}
\crefname{table}{Tab.}{Tabs.}
\crefname{algocf}{Alg.}{Algs.}
\Crefname{algocf}{Algorithm}{Algorithm}

\acrodef{ai}[AI]{Artificial Intelligence}
\acrodef{vict}[ViCT]{visual causal transition model}
\acrodef{wcl}[WCL]{weak-supervised concept learner}
\acrodef{mdp}[MDP]{Markov Decision Process}
\acrodef{scl}[SCL]{Substitution-based Concept Learner}

\author{Yilue Qian$^{1,2,3*}$, Peiyu Yu$^4$, Ying Nian Wu$^4$, Yao Su$^1$, Wei Wang$^1$\textsuperscript{\Letter}, Lifeng Fan$^1$\textsuperscript{\Letter}
\thanks{\textsuperscript{\Letter} Corresponding authors: \texttt{\{wangwei, lifengfan\}@bigai.ai}}
\thanks{* This work was done during Yilue Qian's internship at BIGAI. $^{1}$ National Key Laboratory of General Artificial Intelligence, Beijing Institute for General Artificial Intelligence (BIGAI). $^{2}$ Institute for Artificial Intelligence, Peking University. $^{3}$ Yuanpei College, Peking University.  $^{4}$ Department of Statistics, University of California, Los Angeles (UCLA).}
}

\begin{document}
\maketitle
\thispagestyle{empty}
\pagestyle{empty}

\begin{abstract}
Visual planning simulates how humans make decisions to achieve desired goals in the form of searching for visual causal transitions between an initial visual state and a final visual goal state. It has become increasingly important in egocentric vision with its advantages in guiding agents to perform daily tasks in complex environments. In this paper, we propose an interpretable and generalizable visual planning framework consisting of i) a novel Substitution-based Concept Learner (SCL) that abstracts visual inputs into disentangled concept representations, ii) symbol abstraction and reasoning that performs task planning via the learned symbols, and iii) a Visual Causal Transition model (ViCT) that grounds visual causal transitions to semantically similar real-world actions. Given an initial state, we perform goal-conditioned visual planning with a symbolic reasoning method fueled by the learned representations and causal transitions to reach the goal state. To verify the effectiveness of the proposed model, we collect a large-scale visual planning dataset based on AI2-THOR, dubbed as \textit{CCTP}. Extensive experiments on this challenging dataset demonstrate the superior performance of our method in visual planning. Empirically, we show that our framework can generalize to unseen task trajectories, unseen object categories, and real-world data. Further details of this work are provided at \url{https://fqyqc.github.io/ConTranPlan/}.
\end{abstract}

\section{Introduction}
\label{sec:intro}
\setstretch{0.96}
As one of the fundamental abilities of human intelligence, planning is the process of insightfully proposing a sequence of actions to achieve desired goals, which requires the capacity to think ahead, to employ knowledge of causality, and the capacity of imagination~\cite{walker201322}, so as to reason and foresee the proper actions and their consequences on the states for all the intermediate transition steps before finally reaching the goal state. Visual planning simulates this thinking process of sequential causal imagination in the form of searching for visual transitions between an initial visual state and a final visual goal state. With its advantages in guiding agents to perform daily tasks in the first-person view, visual planning has become more and more important in egocentric vision~\cite{gupta2017cognitive} and embodied AI. In robotics, visual planning could also avoid manually designing the required specific goal conditions, action preconditions, and effects for robots.

Previous works for visual planning can be roughly categorized into three tracks, \ie, neural-network-based models~\cite{sun2022plate, oh2015action}, reinforcement-learning-based models~\cite{rybkin2021model, ebert2018visual} and classic search-based models~\cite{paxton2019visual, liu2020hallucinative}. Neural-network-based models can be trained in an end-to-end manner, which tends to fall short in terms of its interpretability~\cite{gao2023interpretability}. Reinforcement-learning-based models can perform goal-conditioned decisions, but could suffer from sparse reward, low data efficiency~\cite{ladosz2022exploration}, and low environment and task generalization ability~\cite{packer2018assessing}. Considering these limitations and inspired by human cognition, our method falls into the third ``search-based'' category and further proposes three key components for visual planning, namely \textbf{representation learning, symbolic reasoning, and causal transition modeling}. Representation learning focuses on extracting objects' dynamic and goal-oriented attributes. Symbolic reasoning performs action planning at the abstract higher level via learned symbols. Causal transition models the visual preconditions and action effects on attribute changes.

\begin{figure}
   \includegraphics[width=\linewidth,trim=4cm 0.5cm 4cm 0.5cm,clip]{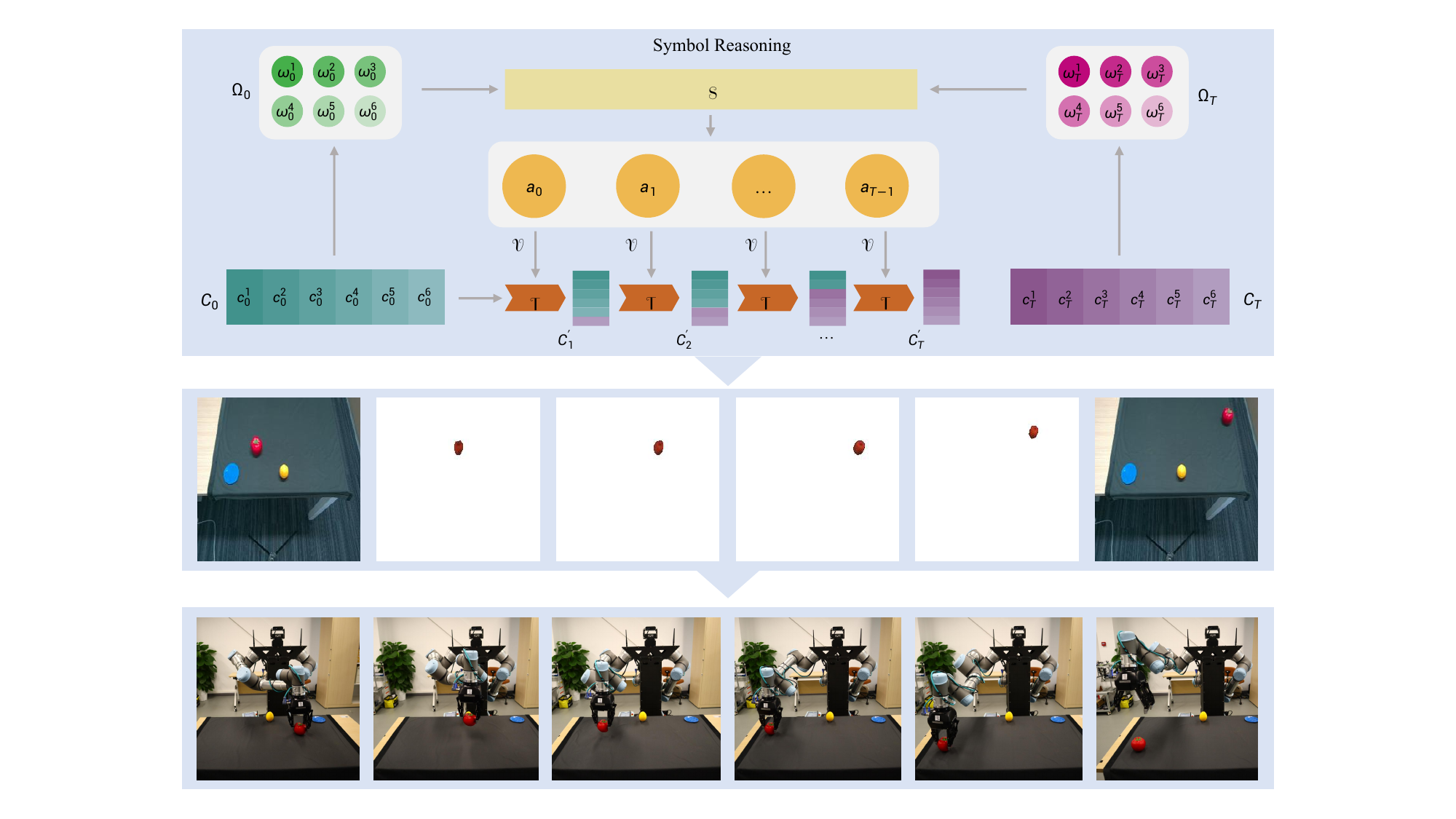}
   \caption{\textbf{Our visual planning framework.} Given an initial state and a goal state, we aim to predict the intermediate states (in the second row) that will guide a robot to manipulate the target objects (in the third row). The disentangled concept-based representation $C$ and abstracted symbol representation $\Omega$, as well as their corresponding causal transition $\mathcal{T}$ and symbol reasoning $S$, are effectively combined into a bi-level planning framework for better generalization (in the first row).
   }
    \label{fig:motiv}
\end{figure}

At the \textbf{perception} level, we propose to learn concept-based disentangled representation and believe such human-like perception ability to abstract concepts from observations is vital for visual causal transition modeling~\cite{zhu2020dark}. The reason is that such representation learning could encode images at a higher semantic level beyond pixels, identifying distinct attribute concepts and isolating ``essential'' factors of variation, thus serving causal learning~\cite{trauble2021disentangled}. This also enhances both robustness and interpretability~\cite{suter2018interventional, trauble2021disentangled, adel2018discovering}, and facilitates compositional generalization to unseen scenarios in zero-shot inference~\cite{atzmon2020causal, trauble2021disentangled, higgins2017darla, locatello2020weakly}, thereby supporting a wide range of real-world downstream tasks. At the \textbf{reasoning and planning} level, we argue that understanding the atomic causal mechanisms is inevitable for task planning. Leveraging learned disentangled representations of concepts, we gain insight into the core of atomic causal transitions, which involves identifying key relevant variable concepts and predicting the outcomes of actions executed upon them. The understanding and reasoning of the abstract higher-level task planning composed of the lower-level atomic causal transition also have the potential to be more generalizable and interpretable~\cite{edmonds2021learning, scholkopf2022causality}. Thus, we propose a \ac{vict} and its abstracted symbolic transition model, which corresponds to the discrete higher-level task planning and avoids the problem of ``error accumulation''~\cite{garcez2022neural}.
Guided by symbolic transition, the visual transition reconstructs intermediate and final goal images. 



Technically, there are \textbf{three critical modules} in our visual planning framework. First, a novel \ac{scl} (\cref{subsec: wcl}) is learned by switching the latent concept representations of a pair of images with different attribute concepts. Second, a set of state symbols is abstracted from clustering low-level concept token representations (\cref{subsec:symbol}). The most efficient symbolic transition path can be found via a Markov Decision Process (MDP). Third, a visual transition model (\cref{sec:vct}) is proposed to learn the action-induced transition of the changeable attributes given the concept representations of the precondition image and thus generate the resulting effect image. To verify the effectiveness of our framework, we collect a large-scale visual planning dataset, which contains a concept learning dataset and a causal planning dataset. Extensive comparison experiments and ablation studies on this dataset demonstrate that our model achieves superior performance in the visual planning task and various forms of generalization tests. 


To summarize our \textbf{main contributions}: (i) We propose a novel concept-based visual planning framework, which models both discrete symbolic transition and continuous visual transition for efficient path search and intermediate image generations. Comprehensive experiments show that our method achieves superior performances in visual task planning and generalization tests. (ii) Apart from generalizability, our method has better interpretability by generating a causal chain (the action sequences and the intermediate state images) to explicitly demonstrate the transition process to the goal. (iii) We collect a large-scale visual planning dataset, which can foster future research in the community.

\subsection{Related Work}
\label{sec:related_work}

\textbf{Visual planning} is feasible with the learned representation and atomic causal effects. \cite{lin2022diffskill} proposed a method for long-horizon deformable object manipulation tasks from sensory observations, which relies heavily on differentiable physics simulators.
\cite{paxton2019visual} performed a tree-search-based planning algorithm on the learned world representation after applying high-level actions for visual robot task planning, but they ignored learning disentangled representations. \cite{sun2022plate} learned how to plan a goal-directed decision-making procedure from real-world videos, leveraging the structured and plannable latent state and action spaces learned from human instructional videos, but their transformer-based end-to-end model is hard to generalize to unseen planning tasks. \cite{oh2015action} proposed a model based on deep neural networks consisting of encoding, action-conditional transformation, and decoding for video prediction in Atari Games, but they do not abstract symbols for efficient reasoning. \cite{silver2021iros} is the most similar to ours, which learned symbolic operators for task and motion planning, but cannot generate intermediate images.

\textbf{Concept-based disentangled representation learning} has emerged as a popular way of extracting human-interpretable representations~\cite{kazhdan2021disentanglement}. Discrete and semantically-grounded representation is argued to be helpful for human understanding and abstract visual reasoning, enables few-shot or zero-shot learning and leads to better down-stream task performance~\cite{van2019disentangled, yu2022latent}. Previous studies tried to learn disentangled concept representation either in a completely unsupervised manner~\cite{chen2016infogan, zhu2020s3vae, higgins2016early, yang2022visual 
}, or via weak supervision and implicit prototype representations~\cite{stammer2022interactive}, or by employing supervision from the linguistic space~\cite{saini2022disentangling, mao2019neuro}. There have been diverse learning techniques, such as Transformer~\cite{yang2022visual}, (sequential) variational autoencoder~\cite{zhu2020s3vae, higgins2016early}, and information maximizing generative adversarial nets~\cite{chen2016infogan}, etc. 
Existing techniques have proved successful on objects mostly with limited variation, such as digits, simple geometric objects~\cite{stammer2022interactive}, and faces~\cite{chen2016infogan}.
In this work, we propose a variant of~\cite{yang2022visual} by imposing more reconstruction constraints, which works very well on more complex household objects with diverse variations (\cref{sec:dataset}) and better benefits the downstream planning task compared to prior works. 

\textbf{Causal reasoning} for task understanding is one of the essential capabilities of human intelligence, and a big challenge for \acs{ai} with the difficulty of generating a detailed understanding of situated actions, their dependencies, and causal effects on object states~\cite{jia2022egotaskqa}. 
Various evaluated state-of-the-art models only thrive on the perception-based descriptive task, but perform poorly on the causal tasks (\ie, explanatory, predictive, and counterfactual tasks), suggesting that a principled approach for causal reasoning should incorporate not only disentangled and semantically grounded visual perception, but also the underlying hierarchical causal relations and dynamics~\cite{yi2019clevrer}. \cite{fire2017inferring} built a sequential Causal And-Or Graph (C-AOG) to represent actions and their effects on objects over time, but suffers from ambiguity in real-life images due to their not-well-disentangled representation. Our work benefits from our disentangled concept representation by finding a latent space where important factors could be isolated from other confounding factors~\cite{atzmon2020causal}, and we ground actions to their causal effects on relevant object attributes. Our bi-level causal planning framework with discrete symbolic transition and continuous visual transition also helps to resist the real-world data noises and ambiguity.

\section{Environment \& Dataset}
\label{sec:dataset}

To facilitate the learning and evaluation of the concept-based visual planning task, we collect a large-scale RGB-D image sequence dataset named \textit{CCTP} (Concept-based Causal Transition Planning) based on AI2-THOR simulator~\cite{ai2thor}. We exclude scene transitions in each task by design to focus more on concept and causal transition learning, \ie, each task is performed on a fixed workbench, although the workbenches and scenes vary from task to task. The frame resolution is $384 \times 256$, converted into $256 \times 256$ at the beginning of our method. 
The whole dataset consists of a concept learning dataset and a visual causal planning dataset, which we will illustrate in detail below.

\subsection{Concept Learning Dataset}

\label{subsec:concept dataset}
We learn six different kinds of concepts: \texttt{TYPE},  \texttt{POSITION\_X}, \texttt{POSITION\_Y}, \texttt{ROTATION}, \texttt{COLOR}, and \texttt{SIZE}. \texttt{TYPE} refers to the object category. The dataset has eight different types of objects in total, including \textit{Bread}, \textit{Cup}, \textit{Egg}, \textit{Lettuce},  \textit{Plate}, \textit{Tomato}, \textit{Pot}, and \textit{Dyer}, all of which can be manipulated on the workbench. We manually add the \texttt{COLOR} concept to the target object by editing the color of the object in its HSV space. This leads to 6 different colors for each object, and $20$ samples are provided for each color to avoid sample bias. For \texttt{SIZE} concept, we rescale each target object to 4 different sizes as its concept set. As for the position, we use \texttt{POSITION\_X} and \texttt{POSITION\_Y} to refer to the coordinates along the horizontal X-axis and the vertical Y-axis w.r.t. the workbench surface. We discretize \texttt{POSITION\_X} with 3 values and \texttt{POSITION\_Y} with 5. Notably, changes in \texttt{POSITION\_X} and \texttt{POSITION\_Y} also cause variant perspectives of an object. For \texttt{ROTATION}, we set $0$, $90$, $180$ and $270$ degrees for all types of objects.
We exhaustively generate all possible target objects with different value combinations of the six concepts, resulting in 234,400 images. Leveraging the masks provided by AI2-THOR, we isolate the foreground images, containing only the target object with a black background. We randomly choose $40\%$ of the concept combinations for training. For each image $X_{0,f}$ in the training set and each concept index $i$, we search for image $X_{1,f}$ within the training set such that $X_{0,f}$ and $X_{1,f}$ differ only in the $i$-th concept. We use such paired images and the corresponding label $i$ for concept learning.

\subsection{Causal Planning Dataset}

\label{subsec:planning dataset}

A causal planning task consists of several steps of state transitions, each caused by an atomic action. We define seven different atomic actions in our dataset, including  \texttt{move\_front}, \texttt{move\_back}, \texttt{move\_left}, \texttt{move\_right}, \texttt{rotate\_left}, \texttt{rotate\_right}, and \texttt{change\_color}. The magnitude of each action is fixed. The target object states (\eg, its color) are randomly initialized in each task from our dataset. The task lengths (\ie, the number of steps for each task) are not fixed.
We collect four subsets of tasks, each representing a difficulty level. In the first level, the workbench has no obstacles, and the ground truth actions involve only movements. In the second level, several fixed obstacles appear on the workbench. In the third level, a dyer additionally appears on the workbench, and the target object must be moved adjacent to the dyer to change its color if necessary before moving to the target position. In the fourth level, rotation actions are involved additionally. The action sequence in each task is paired with the corresponding visual observations. The maximum task lengths for each subset are 6, 9, 15, and 16, and the average are 2.67, 2.81, 4.66, and 5.64, respectively.
Each subset contains 10,000 tasks: 8,000 for training, 1,000 for validation, and 1,000 for testing.

We construct additional generalization test benchmarks based on our collection. We provide four levels of \textbf{Unseen Object} generalization tests for object-level generalization. For each level, we generate 1000 tasks with the target object types unseen in the training dataset, including object types of \textit{Cellphone}, \textit{Dish Sponge}, \textit{Saltshaker}, and \textit{Potato}. Additionally, we provide datasets for generalization tests for unseen tasks. The training and testing tasks in the \textbf{Unseen Task} dataset have different combinations of action types. For example, the training dataset may include tasks that consist of only \texttt{move\_left} and \texttt{move\_front} actions, as well as tasks that consist of only \texttt{move\_right} and \texttt{move\_back} actions, while the testing dataset contains tasks from the held-out data with different combinations. The \textbf{Unseen Task} dataset is limited to level-1 and level-2 because limited combinations of actions are insufficient to accomplish more difficult tasks. 

\begin{figure}[t!]
   \includegraphics[width=\linewidth,trim=10cm 19cm 19cm 17cm,clip]{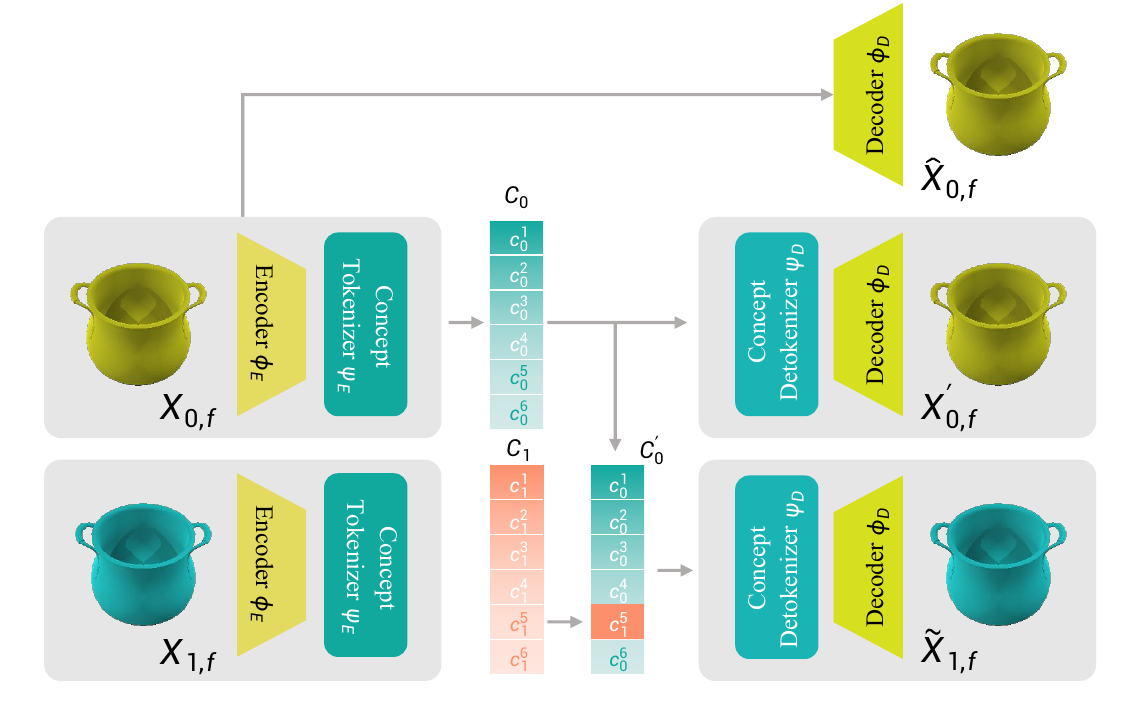}
   \caption{\textbf{Architecture of SCL}. Foreground images $X_{0,f}$ and $X_{1,f}$ differ only in the \texttt{COLOR} concept. After extracting their concept tokens and assuming the token $c^5_i$ to represent the color concept, the \texttt{COLOR} concept $c_0^5$ of $X_{0,f}$ is substituted by $c_1^5$ from $X_{1,f}$, which are then fed into the detokenizer and decoder to reconstruct images.}
\label{fig:concept learning}
\end{figure}

\section{Method}
\label{sec: model}

Given an initial RGB-D state image $X_0$ and a final RGB-D state image $X_T$, our task is to find a valid and efficient state transition path with an inferred sequence of actions $\bm{\Gamma}=\{a_t\}_{t=1,..., T}$, as well as generating intermediate and final state images $\Tilde{\bm{X}}=\{\Tilde{X_t}\}_{t=1,..., T}$. To fulfill this task, we use a concept learner to extract disentangled concept representations for state images, abstract concept symbols for reasoning, and train a \ac{vict} to generate intermediate state images.


\begin{figure*}[t!]
\centering
   \includegraphics[width=\linewidth,trim=5.1cm 21cm 5.1cm 20cm,clip]{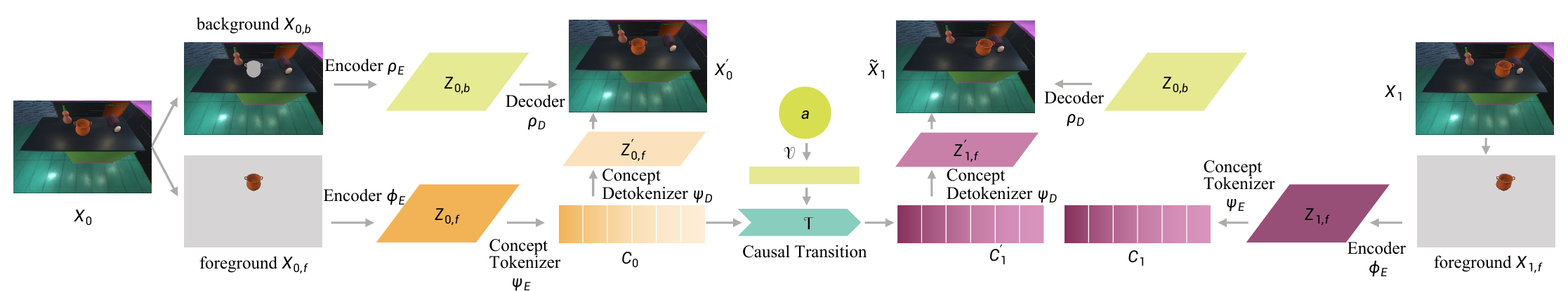}
   \caption{\textbf{Architecture of ViCT}. The concept tokenizer extracts object concept tokens for causal transition. The causal transition model transforms concept tokens from $C_0$ to $C_1^{'}$ with the action embedding $\mathcal{V}(a)$. The background encoder converts the background image into latent vectors, which are then combined with predicted concept tokens $C_1^{'}$ to generate the effect image $\Tilde X_1$.}
\label{fig:causal transition learning}
\end{figure*}

\subsection{Substitution-based Concept Learner}
\label{subsec: wcl}


The architecture of our \ac{scl} is illustrated in \cref{fig:concept learning}. A pair of foreground images $X_{0,f}$ and $X_{1,f}$ are given as input, where these two images contain two objects differing only in one concept, \eg, a yellow pot and a green pot. Then a shared encoder $\mathcal{\phi}_E$ is applied to the foreground images to obtain the latent embeddings $Z_{i,f}=\mathcal{\phi}_E(X_{i,f})$. The embedding $Z_{i,f}$ is further fed into a concept tokenizer $\mathcal{\psi}_T$ to generate the concept tokens $C_i=\{c^k_i\}_{k=1,...,6}=\mathcal{\psi}_T(Z_{i,f})$. Here $k$ is the concept index, and we assume there exist six visual concepts, \ie, \texttt{TYPE},  \texttt{POSITION\_X}, \texttt{POSITION\_Y}, \texttt{ROTATION}, \texttt{COLOR}, and \texttt{SIZE}, representing the visual attributes of the target objects (refer to \cref{subsec:concept dataset} for details).



The concept token $c^i_0$ is substituted with $c^i_1$ to get a new concept token vector $C_0^{'}$, where $i$ indexes the different concept between the paired images $X_{0,f}$ and $X_{1,f}$. For example, the token $c^5_i$ assumes to represent the \texttt{color} concept in \cref{fig:concept learning}, so replacing $c_0^5$ with $c_1^5$ will change the original yellow pot to a green pot. The token vector $C_0^{'}$ is fed into a concept detokenizer $\mathcal{\psi}_D$ to reconstruct the latent embedding $Z_{1,f}^{'}=\mathcal{\psi}_D(C_0^{'})$, which is further decoded into image $\Tilde X_{1,f}=\mathcal{\phi}_D(Z_{1,f}^{'})$. After the concept detokenizer and decoder, we obtain a combined reconstruction loss as follows:
\begin{equation}
\small
    \mathcal{L}_1 =  \mathcal{L}_{MSE}(X_{0,f}^{'}, X_{0,f})+ \mathcal{L}_{MSE}(\Tilde X_{1,f}, X_{1,f}),
    \label{eq:concept loss_1}
\end{equation}
where $\mathcal{L}_{MSE}$ is the mean squared error.
In addition, we add another branch that directly connected the encoder to the decoder. This branch aims to distinguish the role of the encoder from that of the concept tokenizer; it enforces the encoder to learn hidden representations by reconstructing $X_{0,f}$. The reconstructed image and reconstruction loss of this branch are $\hat{X}_{0,f}$ and $\mathcal{L}_{MSE}(\hat{X}_{0,f}, X_{0,f})$, respectively.
Similar to~\cite{yang2022visual}, a Concept Disentangling Loss (CDL) is employed to reduce interference between the concept tokens. The CDL can be formulated as follows:
\begin{equation}
\small
    \mathcal{L}_{CDL} =  \mathcal{L}_{CE}(\lVert C_0-C_1 \rVert_2, i),
    \label{eq:concept loss_2}
\end{equation}
where $\mathcal{L}_{CE}$ is the cross-entropy loss. $\lVert C_0-C_1 \rVert_2$ calculates the $l_2$ norm of the variation of each concept token. $i$ is the ground-truth token index and indicates that the $i$-th concept token is replaced. The total loss $\mathcal{L}_C$ of concept learner is as follows:
\begin{equation}
\small
    \mathcal{L}_C = \mathcal{L}_1+ \mathcal{L}_{MSE}(\hat{X}_{0,f}, X_{0,f})
    + \mathcal{L}_{CDL},
    \label{eq:concept loss}
\end{equation}
where the equal weights for each loss work well in our experimental settings.


\subsection{Symbol Abstraction and Reasoning}
\label{subsec:symbol}

Symbol abstraction aims to convert concept tokens into discrete symbols for later symbolic reasoning. Our empirical results in \cref{fig:action concept eff} show that the concept tokens learned in \cref{subsec: wcl} are well-disentangled and can be easily clustered into several categories. Therefore, a clustering algorithm could be applied to the concept tokens to generate symbols. 
Specifically, we collect all the concept tokens extracted from the training data using the \ac{scl} and create the concept token spaces: $\mathbf{C}=\left\{c_n\right\}$.
Then, we employ the K-means algorithm to cluster data points within the concept spaces, resulting in the concept centers $\left\{\bar c\right\}$ and a symbol assignment $\omega = \sigma(c, \left\{\bar c\right\})$ for each concept token $c$. Here, $\sigma$ is the nearest neighbor function, which assigns the symbol of the nearest concept center to $c$. This procedure is independently applied to six defined concepts, with each concept assigned a specific number of clustering centers that correspond to their predefined value spaces, abstracting a set of concept symbols $\Omega=\left\{\omega^k\right\}_{k=1,\dots,6}$ for each image. 


The symbolic reasoning aims to find the most plausible transition path from the initial state to the goal state at the symbol level, which can be formulated as an MDP. 
Given the initial concept symbols $\Omega_0=\{\omega_0^k\}_{k=1,\dots,6}$ and the action $a_0$, the symbol reasoner computes the distribution of concept symbols at the next timestep $\Pr\left[\Omega_1^{'}\mid a_0, \Omega_0\right]$.  
The concept symbol distribution at the timestep $t$ can be obtained as follows:
\begin{equation}
\label{eq:mdpsigma1}
\small
\begin{split}
    &\Pr\left[{\Omega}_t^{'}\mid a_{0:t-1}, \Omega_0\right]\\
    =&\sum_{o\in\mathbf{\Omega}}\Pr\left[\Omega_t^{'}\mid a_{t-1},\Omega_{t-1}^{'}=o\right]\cdot\Pr\left[\Omega_{t-1}^{'}=o\mid a_{0:t-2},\Omega_0\right],   
\end{split}
\end{equation}
where $\mathbf{\Omega}$ denotes the the entire concept symbol space.
Additionally, two legality checks are implemented during the reasoning process to ensure the validity of the action sequence, involving action legality and state legality checks. The action legality is defined as ${\pmb{1}}_{\Pr\left[a\mid \Omega\right]> {\rm thresh}}$.
This check aims to prevent the use of noise-inducing transformations caused by the \ac{scl}, thereby modifying \cref{eq:mdpsigma1} to: 
\begin{equation}
\small
\label{eq:mdpsigma2}
\resizebox{0.94\linewidth}{!}{
$\begin{split}
        &\Pr\left[{\Omega}_t^{'}\mid a_{0:t-1}, \Omega_0\right]\\
        =&\sum_{o\in\mathbf{\Omega}}{\pmb{1}}_{\Pr\left[a_{t-1}\mid o\right]> {\rm thresh}}\Pr\left[\Omega_t^{'}\mid a_{t-1},\Omega_{t-1}^{'}=o\right]\Pr\left[\Omega_{t-1}^{'}=o\mid a_{0:t-2},\Omega_0\right].
\end{split}$}
\end{equation}
The state legality check is designed to eliminate contributions to the distribution originating from invalid states (e.g., collisions with obstacles on the workbench). It can be written as follows:
\begin{equation}
\small
\begin{split}
    &\Pr\left[{\Omega}_t^{'}=o_0\mid a_{0:t-1}, \Omega_0;\{\Omega_{\rm env}\}\right]\\=&\frac{\pmb{1}_{o_0\in\mathbf{\Omega}_{\rm valid}}\cdot\Pr\left[{\Omega}_t^{'}=o_0\mid a_{0:t-1}, \Omega_0\right]}{\sum_{o\in\mathbf{\Omega}_{\rm valid}}\Pr\left[{\Omega}_t^{'}=o\mid a_{0:t-1}, \Omega_0\right]}
\end{split}
\end{equation}
where $\mathbf{\Omega}_{\rm valid}\subseteq \mathbf{\Omega}$ represents the set of valid concept symbols given the concept symbols of other objects in the environment, and $o_0$ is an arbitrary element within $\mathbf{\Omega}$. 
To reduce computational complexity, the reasoning process is individually applied to each concept. This approach is effective due to the well-designed disentangled concepts, which ensure that the changes in each concept are independent given a particular action. 
The MDP estimates symbol-level transition probability distribution by recording the (input, action, output) triplets in the training data. 
The objective is to discover the action sequence $a_{0:T-1}$ that is most likely to result in a distribution of concept symbols ${\Omega}_T^{'}$ closely approximating the goal concept symbols $\Omega_T$. This action sequence is then passed into the \ac{vict} (See \cref{sec:vct}) to generate predicted intermediate images (\cref{fig:motiv}).

\subsection{Visual Causal Transition Learning}

\label{sec:vct}

The aim of \ac{vict} is to generate visual effect images based on precondition images and human actions. For example, \cref{fig:causal transition learning} shows an action that moves the pot one step to the right. \ac{vict} predicts image $\Tilde X_1$ by transforming the the pot in image $X_0$ with a \texttt{move\_right} action.

As seen in \cref{fig:causal transition learning}, three parts exist in the framework of \ac{vict}.
Firstly, the causal transition is the key part of \ac{vict}. This process transforms object concept tokens from $C_0$ to $C_1^{'}$ with the help of an action embedding $\mathcal{V}(a)$. The action $a$ is encoded into a one-hot vector and further embedded via an embedding function $\mathcal{V}$ to achieve this. The transition process is as follows:
\begin{equation}
\small
    C_1^{'}=\mathcal{T}(C_0, \mathcal{V}(a)),
\label{eq:transition model}
\end{equation}
where $C_1^{'}$ represents the resulting concept tokens. $\mathcal{T}$ denotes the causal transition function involved in this process.
In addition to the causal transition component, two other crucial parts in \ac{vict} are dedicated to managing visual extraction and reconstruction.
The second part contains a concept tokenizer to extract foreground object concept tokens $C_0$ for later transitions. This concept tokenizer has been trained as described in \cref{subsec: wcl} and fixed here. This part also involves a background encoder $\rho_E$, which processes the background image to produce latent vectors represented as $Z_{0,b}$.
The vectors $Z_{0,b}$ store background-related information and will be used to generate the resultant image $\Tilde{X}_1$. 
The third part combines foreground object concept tokens and background latent vectors to predict effect image $\Tilde{X}_1$ with the background decoder $\rho_D$. Instead of directly using concept tokens, we convert them back to latent embeddings, \ie, from $C_1^{'}$ to $Z_{1,f}^{'}$, and then concatenate $Z_{1,f}^{'}$ with latent vectors $Z_{0,b}$ as the input to the decoder. Similarly, we can also combine $Z_{0,f}^{'}$ and $Z_{0,b}$ to obtain a reconstruction image $X_0^{'}$.

Up to now, two losses can be computed during training: a reconstruction loss $\mathcal{L}_{MSE}(X_0^{'}, X_0)$ and a prediction loss $\mathcal{L}_{MSE}(\Tilde{X}_1, X_1)$. In addition to measuring image-level prediction errors, we can also evaluate token-level prediction errors. Given a ground-truth effect image $X_1$, we extract its concept tokens $C_1$, and introduce a token prediction loss $\mathcal{L}_{MSE}(C_1^{'}, C_1)$.
The total loss of \ac{vict} is summarized as follows:
\begin{equation}
\small
    \mathcal{L}_T=\mathcal{L}_{MSE}(C_1^{'}, C_1)+ \mathcal{L}_{MSE}(\Tilde{X}_1, X_1) + \mathcal{L}_{MSE}(X_0^{'}, X_0).
    \label{eq:loss L_T}
\end{equation}
The \ac{vict} is trained on our causal planning dataset (see \cref{subsec:planning dataset}).

\section{Experiments}

\label{sec:result}

Our experiments aim to answer the following questions: (1) Is our model design effective and applicable to visual planning tasks? (2) How do the proposed key components contribute to the model performance? (3) Are the learned concepts and causal transitions interpretable? (4) Does the proposed method exhibit generalization on novel tasks? To answer these questions, we perform extensive experiments, showing the proposed methods are interpretable, generalizable, and capable of producing significantly better results than baseline methods.



\subsection{Evaluating Visual Planning on Dataset \textit{CCTP}}

\begin{figure*}[t!]
\centering
   \includegraphics[width=\linewidth,trim=2cm 2cm 2cm 2cm,clip]{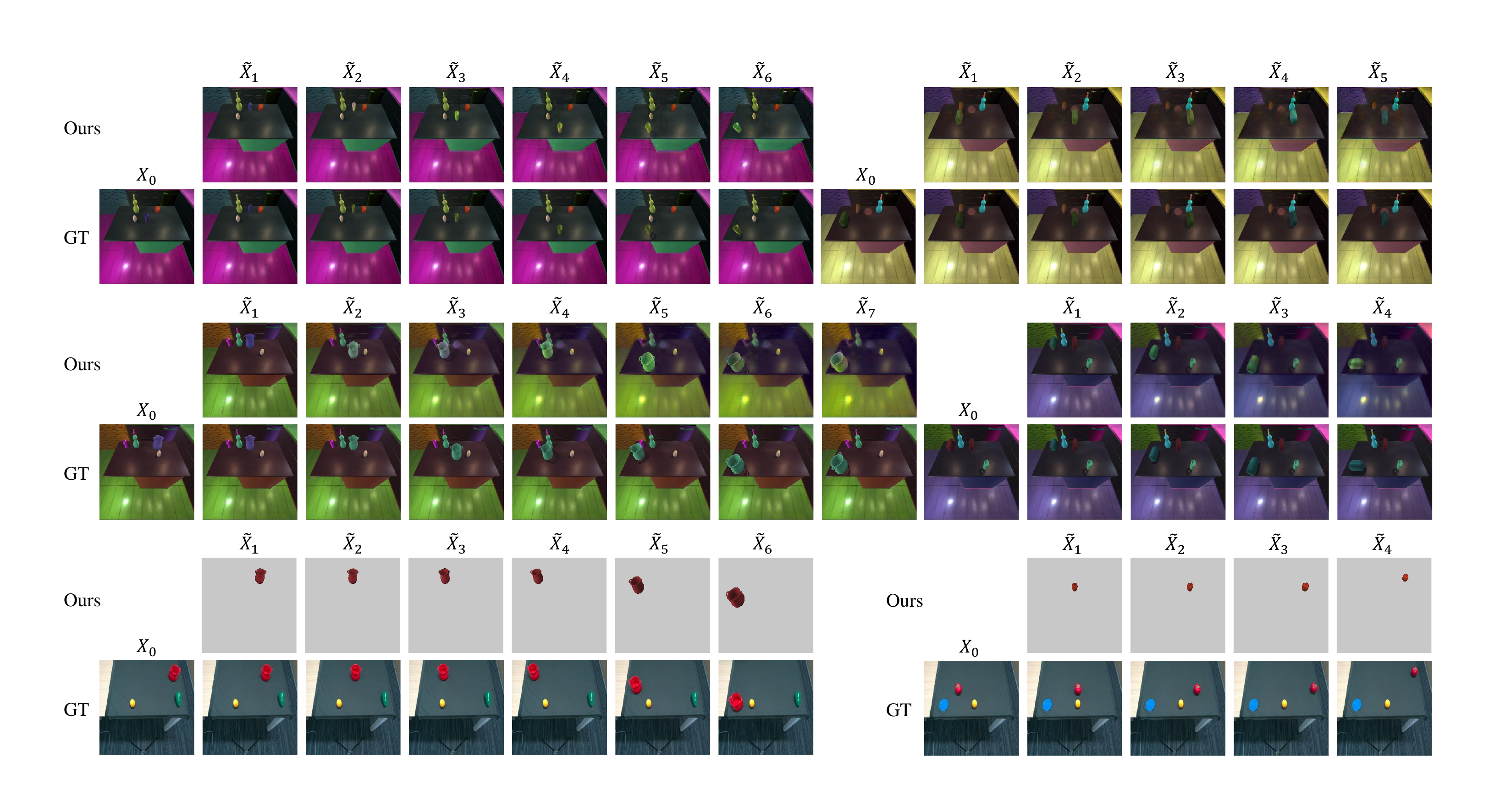}
   \caption{{\bf Qualitative results of our visual planning model}. The top two samples are obtained from the level-3 dataset, and the middle two are from the level-4 dataset. Our model demonstrates its ability to manage tasks of varying lengths, effectively plan action sequences, and generate intermediate and goal state images. Notably, the first sample from the level-4 dataset generates a different path than the ground truth but still achieves success and maintains high efficiency. The bottom two samples are from our real-world data experiments, corresponding to the level-3 dataset. To simplify implementation, we focus on visually planning the target objects in the real-world images, ignoring the background.}
\label{fig:plan exp}
\end{figure*}

To validate the effectiveness of our model design, we employ PlaTe~\cite{sun2022plate}, the state-of-the-art method for visually-grounded planning, as our baseline. To probe the contribution of our proposed components, we replace each component with alternative baselines to compare with. We replace the proposed concept learner with strong baselines such as beta-VAE~\cite{higgins2016beta} and VCT~\cite{yang2022visual} model to verify the effectiveness of our concept learning module. Additionally, we compare our model to a goal-conditioned Double DQN agent
~\cite{van2016deep}
trained with prioritized experience replay~\cite{schaul2015prioritized}, noted as ``w/ RL''.
Furthermore, to verify the necessity of our symbolization process, we apply the reasoning process directly to the concept tokens, employing our causal transition model to search for states closest to the goal state within the concept token spaces. We also conduct experiments where we further remove the concept learning process. Instead, we use an autoencoder to extract latent embedding for causal transition. The corresponding results are denoted as ``w/o. symbol'' and ``w/o. concept'', respectively. The ``w/o. concept'' experiments are limited to the level-1 dataset, as this method is unable to avoid object collisions in the higher-level datasets.
Finally, we replace the explicit planning module with a transformer architecture. It takes the initial and goal state concept symbols, provided by our concept learner and symbolizer, as inputs to generate the action sequence.  We refer to this variant as ``w/o causal''. We also substitute the planning module with random action predictions for each step as an additional baseline for reference. 


\paragraph{Evaluation metrics} To thoroughly inspect the performance of visual planning, we design metrics including Action Sequence Prediction Accuracy (ASAcc), Action Sequence Efficiency (ASE), and Final State Distance (FSD). ASAcc is measured as the success rate of sequence prediction. In level-1 and level-2 tasks, a successful prediction entails moving the target object accurately to the position of goal states without encountering any collisions with obstacles (if present). In level-3 tasks, when the target object's color changes, success requires moving the object adjacent to the dyer, applying the \texttt{change\_color} action, and then moving it to the goal position. In level-4 tasks, the target object must also be correctly rotated for success. 
During testing, MDP and search-based methods, including ``Ours'', ``Ours w/ $\beta$-VAE'', ``Ours w/o symbol'', and ``Ours w/o concept'' generate the 5 most possible paths, and randomized algorithms ``Chance'' and ``Ours w/ RL'' make 5 attempts for each task. The top-1 accuracy evaluates the success of the most likely path or the initial attempt, while the top-5 accuracy checks if any of the 5 paths are successful.
ASE measures the efficiency of the planning by comparing the length of the ground truth sequence to that of the predicted sequence. It only considers the successfully predicted sequences. The ASE is defined as follows:
\begin{equation}
\small
    ASE  = \frac{\sum_{i=1}^N\mathbb I(\bm{\Gamma}_i^{pred})\ell(\bm{\Gamma}_i^{gt})/\ell(\bm\Gamma^{pred}_i)}{\sum_{i=1}^N\mathbb I(\bm\Gamma_i^{pred})},
    \label{eq:efficiency}
\end{equation}
where $\mathbb I$ is a indicator function for a successful prediction, $\ell$ represents the length of an action sequence. Of note, the ground truth action sequences in \textit{CCTP} are the most efficient, so the efficiency of a predicted sequence will be no more than 1.
FSD calculates the distance between the positions of the foreground object in the final predicted state and in the goal state. The distance is defined based on the object's coordinates w.r.t. the workbench.
\paragraph{Results} 

We can see from \cref{tab:performance} that the proposed method achieves significantly higher performance compared with baselines. Specifically, we compare our method with different ablative variants on \textit{CCTP} dataset. Our method outperforms baselines in terms of ASAcc by a large margin and achieves the smallest FSD, which demonstrates our method can obtain an accurate planning path to reach the goal state. Our method achieves very competitive ASE. Notably, certain baselines (\eg, Ours w/ RL) attain high levels of ASE, but with a disproportionately lower ASAcc.
Moreover, our model maintains strong performance when encountering hard tasks, while competitive baselines' performances significantly decrease as task difficulty increases. These results demonstrate the effectiveness of our model design. Our full model achieves the best overall performance in all four levels of tests, and each component of our model contributes remarkably to the performance improvements. 
The qualitative results are shown in \cref{fig:plan exp}.

\begin{table}[tb!]
    \caption{\textbf{Quantitative results for visual task planning.} 
    The best scores are marked in \textbf{bold}.}
    \centering
        \scalebox{0.59}{
        \begin{tabular}{lcccccccc}
            \toprule
            \multicolumn{1}{c}{\multirow{4}{*}{\makecell[c]{Model\\ID}}} & 
            \multicolumn{2}{c}{ASAcc.(\%)($\uparrow$)} & 
            \multicolumn{1}{c}{\multirow{2}{*}{ASE($\uparrow$)}} &
            \multicolumn{1}{c}{\multirow{2}{*}{FSD($\downarrow$)}} & 
            \multicolumn{2}{c}{ASAcc.(\%)($\uparrow$)} & 
            \multicolumn{1}{c}{\multirow{2}{*}{ASE($\uparrow$)}} &
            \multicolumn{1}{c}{\multirow{2}{*}{FSD($\downarrow$)}} \\

            \cmidrule(r){2-3} \cmidrule(r){6-7}

            \multicolumn{1}{c}{} & 
            \multicolumn{1}{c}{Top-1} & 
            \multicolumn{1}{c}{Top-5} &
            \multicolumn{1}{c}{} & 
            \multicolumn{1}{c}{} &
            \multicolumn{1}{c}{Top-1} & 
            \multicolumn{1}{c}{Top-5} &
            \multicolumn{1}{c}{} & 
            \multicolumn{1}{c}{} \\
            \cmidrule(r){2-5} 
            \cmidrule(r){6-9}

            \multicolumn{1}{c}{} &
            \multicolumn{4}{c}{Dataset level-1} &
            \multicolumn{4}{c}{Dataset level-2} \\
            
            \toprule
            Chance & 1.3 & 7.3 & -  & 3.139 & 0.4 & 2.2 & - & 3.499  \\
            PlaTe~\cite{sun2022plate} & 38.9 & - & - & - & 15.3 & - & - & -  \\
            \midrule
            Ours w/ $\beta$-VAE~\cite{higgins2016beta} & 0.5 & 3.0 & 0.970 & 3.220 & 0.0 & 3.5 & - & 3.670 \\
            Ours w/ VCT~\cite{yang2022visual} & 54.1 & 60.6 & 0.972 &  1.483 & 1.6 & 4.9 & 0.988 & 1.294   \\
            Ours w/o symbol & 65.8 & 76.9& 0.983 & 1.197& 41.0 & 52.6 & 0.962 & 1.627  \\
            Ours w/o concept & 56.9 & 77.6& 0.986 & 1.644 &-&-&-&-\\
            Ours w/o causal & 1.4 & - & - & 3.326  & 0.3 & - & - & 3.419\\
            Ours w/ RL & 29.7 & 35.1 & \textbf{0.991}& 2.418  & 2.5 & 6.0 & \textbf{1.000}&3.150 \\
            \cellcolor{mygray}\textbf{Ours} & \cellcolor{mygray}\textbf{97.9} & \cellcolor{mygray}\textbf{99.2} & \cellcolor{mygray}0.971 & \cellcolor{mygray}\textbf{0.025}& \cellcolor{mygray}\textbf{99.4} & \cellcolor{mygray}\textbf{99.6} & \cellcolor{mygray}0.981 & \cellcolor{mygray}\textbf{0.013}\\
            
            \toprule
            \multicolumn{1}{c}{} &
            \multicolumn{4}{c}{Dataset level-3} &
            \multicolumn{4}{c}{Dataset level-4} \\
            
            \midrule
            Chance & 0.0 & 0.4 & - & 3.513  & 0.1 & 0.4 & - & 3.147  \\
            PlaTe~\cite{sun2022plate} & 0.7 & - & - & -  & 0.4 & - & - & -   \\
            \midrule
            Ours w/ $\beta$-VAE~\cite{higgins2016beta}  & 0.0 & 0.5 & - & 3.596 & 0.0 & 0.0 & - & 3.107  \\
            Ours w/ VCT~\cite{yang2022visual} & 0.7 & 1.2 & 0.968 & 3.442    & 0.2 & 0.3 & 1.000 & 3.193 \\
            Ours w/o symbol &15.4 & 24.1& 0.970& 2.278 & 9.8&14.0 &0.981 & 2.149\\
            Ours w/o causal & 0.0 & - & - & 3.691 & 0.0 & - & - & 3.201\\
            Ours w/ RL & 3.0 & 3.9 & \textbf{1.000} & 3.030 & 2.8 & 3.5 & \textbf{1.000} & 2.498 \\
            \cellcolor{mygray}\textbf{Ours} & \cellcolor{mygray}\textbf{86.5}& \cellcolor{mygray}\textbf{87.0} & \cellcolor{mygray}0.966 & \cellcolor{mygray}\textbf{0.037} &\cellcolor{mygray}\textbf{55.1}& \cellcolor{mygray}\textbf{76.7} & \cellcolor{mygray}0.978 & \cellcolor{mygray}\textbf{0.003}\\
            \bottomrule
        \end{tabular}%
        }
    \label{tab:performance}
    \vspace{-4.2pt}
\end{table}

\subsection{Interpretable Concepts and Causal Transitions}

\begin{figure}[t!]
\centering
   \includegraphics[width=\linewidth]{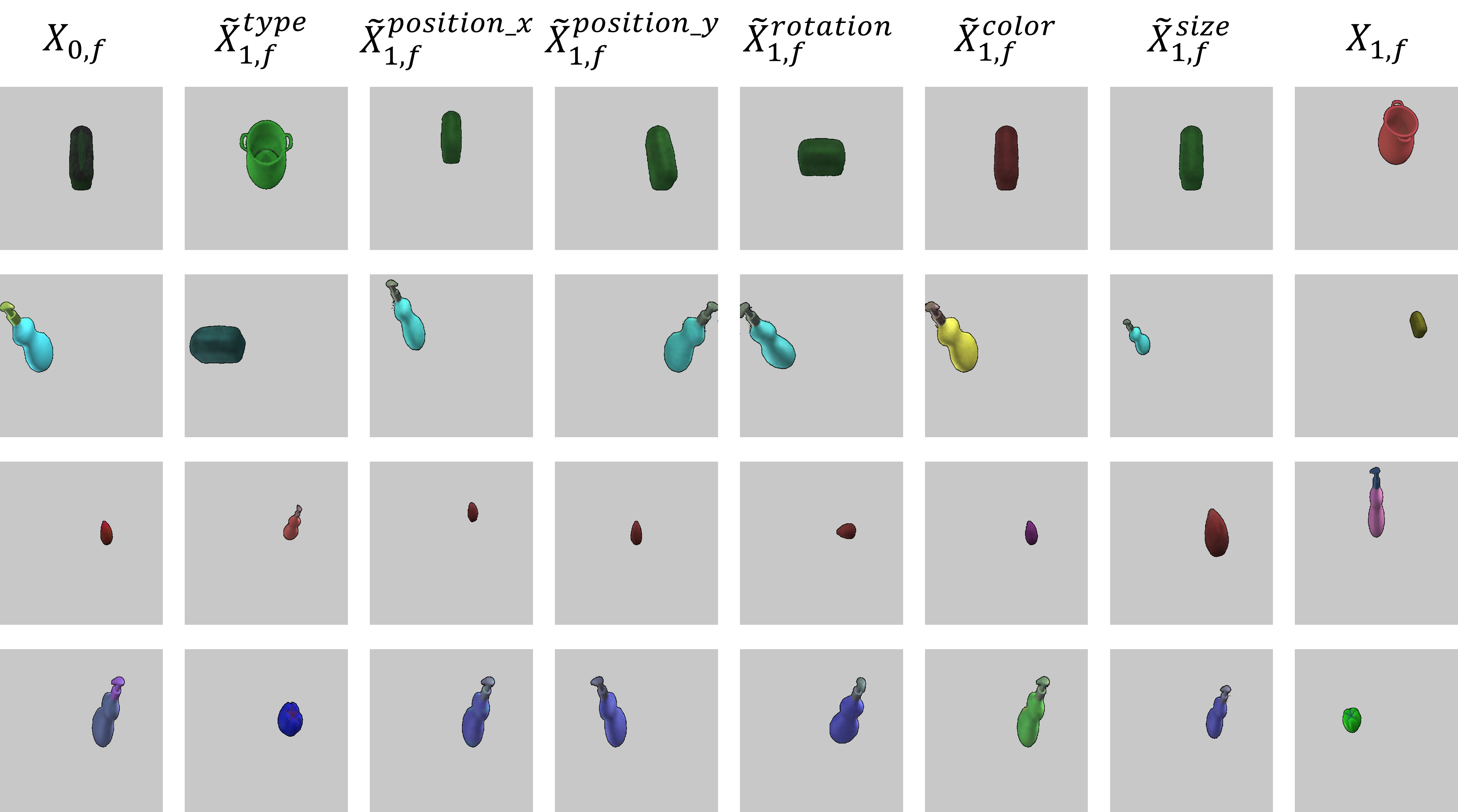}
   \caption{\textbf{Fine-grained attribute-level concept manipulation.} The concept learner generates new images by substituting each concept token $c_0^{i}$ from $X_{0,f}$ with $c_1^{i}$ from $X_{1,f}$.}
\label{fig:concept exp}
\end{figure}

We qualitatively show the interpretability of the concept learned by our model. We randomly choose 2 images $X_{0,f}$ and $X_{1,f}$, substituting the concept token $c_0^{i}$ with $c_1^{i}$ for $i=1,2,3,4,5,6$, which are then fed into the concept detokenizer and the decoder to generate new images. As \cref{fig:concept exp} shows, with the properly learned concept representations, we could perform fine-grained attribute-level concept manipulation. This indicates that our concept learner is capable of disentangling concept factors and demonstrates the interpretability of our method.

\begin{figure}[t!]
\centering
    \includegraphics[width=\linewidth,trim=0.2cm 0cm 0cm 1.6cm,clip]{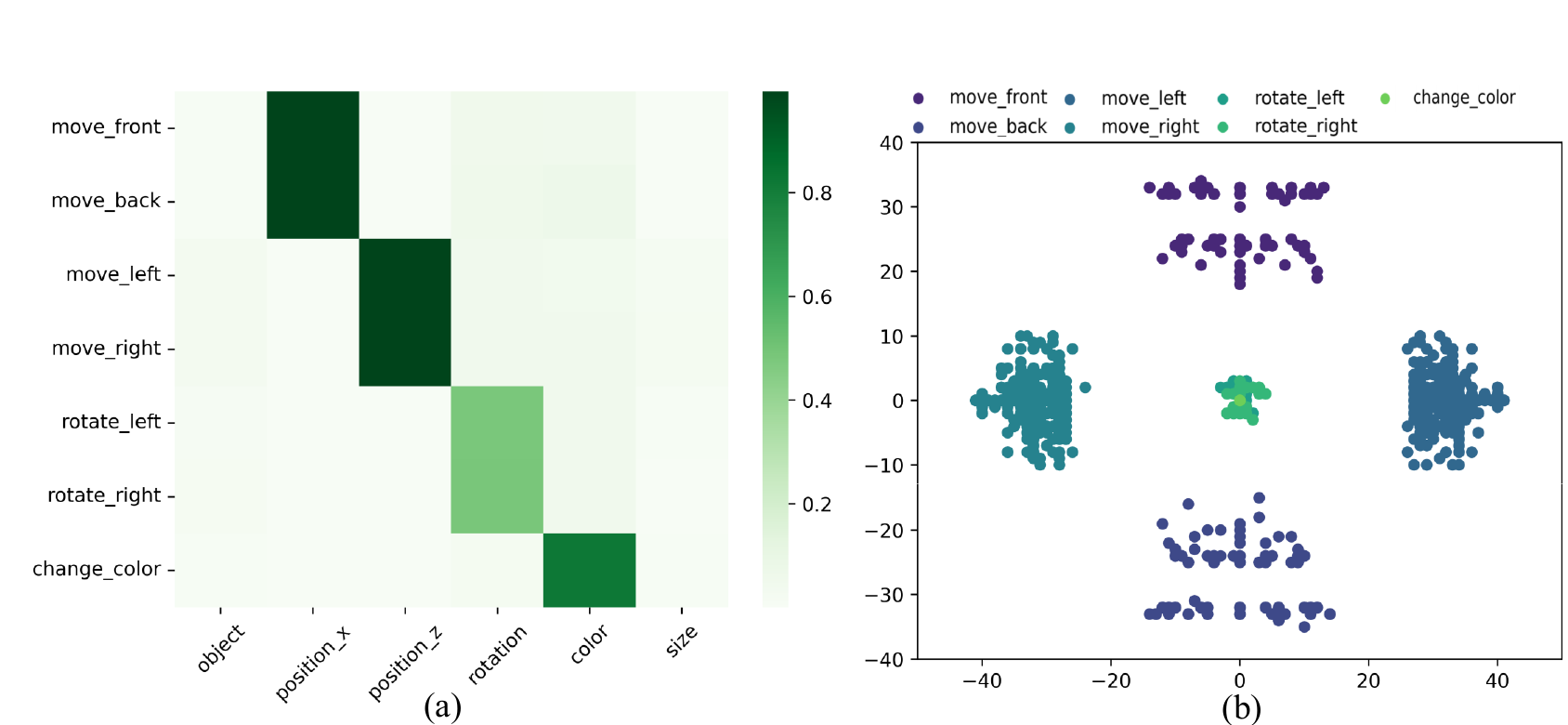}
   \caption{\textbf{Action effects on the learned disentangled concept representations.} (a) $l_2$ norm between the concept vectors before and after each action. (b) Distributions of position change induced by each action.}
\label{fig:action concept eff}
\end{figure}

We quantitatively demonstrate the interpretability of our learned causal transitions with statistics of the corresponding causal effects. To be specific, we aim to answer the question: do the learned causal transitions have semantic meaning consistent with the corresponding action? \cref{fig:action concept eff} (a) shows the correlation between concepts and actions, measured with $l_2$ norm between the concept vectors before and after each action. A larger $l_2$ norm means a higher correlation. We can see that the learned rotation actions only affect the rotation status in the concept vector. Similarly, the horizontal and vertical movements only affect the x and y coordinates. \cref{fig:action concept eff} (b) shows the distribution of position change induced by 7 displacement actions. For example, the position changes of \texttt{move\_front} distribute along the positive y-axis, while those of \texttt{move\_back} distribute along the negative y-axis. 
This evidence indicates that 1) our learned concept is successfully disentangled, which makes it possible for our model to learn causal transitions, and 2) the learned causal transition is consistently grounded to real-world actions with similar semantics.

\subsection{Generalization Tests}

\label{subsec:generalization exp}

\begin{table}[t!]
    \centering
    \caption{\textbf{Quantitative results for generalization tests.} 
    The best scores are marked in \textbf{bold}.}
        \scalebox{0.58}{
        \begin{tabular}{clcccccccc}
            \toprule
            \multirow{23}{*}{\rotatebox[origin=c]{90}{\textbf{Unseen Object}}} &
            \multicolumn{1}{c}{\multirow{4}{*}{\makecell[c]{Model\\ID}}} & 
            \multicolumn{2}{c}{ASAcc.(\%)($\uparrow$)} & 
            \multicolumn{1}{c}{\multirow{2}{*}{ASE($\uparrow$)}} &
            \multicolumn{1}{c}{\multirow{2}{*}{FSD($\downarrow$)}} & 
            \multicolumn{2}{c}{ASAcc.(\%)($\uparrow$)} & 
            \multicolumn{1}{c}{\multirow{2}{*}{ASE($\uparrow$)}} &
            \multicolumn{1}{c}{\multirow{2}{*}{FSD($\downarrow$)}} \\

            \cmidrule(r){3-4} \cmidrule(r){7-8}

            &\multicolumn{1}{c}{} & 
            \multicolumn{1}{c}{Top-1} & 
            \multicolumn{1}{c}{Top-5} &
            \multicolumn{1}{c}{} & 
            \multicolumn{1}{c}{} &
            \multicolumn{1}{c}{Top-1} & 
            \multicolumn{1}{c}{Top-5} &
            \multicolumn{1}{c}{} & 
            \multicolumn{1}{c}{} \\
            \cmidrule(r){3-6} 
            \cmidrule(r){7-10}

            &\multicolumn{1}{c}{} &
            \multicolumn{4}{c}{Dataset level-1} &
            \multicolumn{4}{c}{Dataset level-2} \\
            \toprule
            
            &Chance & 0.6 & 4.7 & - & 3.203  & 1.1 & 3.2 & - & 3.591\\
            &PlaTe~\cite{sun2022plate} & 18.5 & - & - & -& 9.7 & - & - &-\\
            &Ours w/o symbol & 44.0 & 59.9 & 0.968 & 1.507& 29.0 & 43.8 & 0.986 & 1.880 \\
            &Ours w/o concept & 37.1 & 60.5 & 0.950 & 1.319 &-&-&-&-  \\
            &Ours w/o causal & 1.7 & - & - & 3.233  & 0.2 & - & - & 3.563\\
            &Ours w/ RL & 30.2 & 35.9 & \textbf{0.989} & 1.887& 2.2 & 6.1 & \textbf{1.000} & 3.549  \\
            & \cellcolor{mygray}\textbf{Ours} & \cellcolor{mygray}\textbf{72.4} & \cellcolor{mygray}\textbf{97.2} & \cellcolor{mygray}0.987 & \cellcolor{mygray}\textbf{0.470} &\cellcolor{mygray}\textbf{73.2} & \cellcolor{mygray}\textbf{93.6} & \cellcolor{mygray}0.978 & \cellcolor{mygray}\textbf{0.491}  \\

            \cmidrule{2-10} 
            &\multicolumn{1}{c}{} &
            \multicolumn{4}{c}{Dataset level-3} &
            \multicolumn{4}{c}{Dataset level-4} \\
            
            \cmidrule{2-10} 
            & Chance & 0.0 & 0.0 & - & 3.544  & 0 & 0.1 & - & 3.518\\
            & PlaTe~\cite{sun2022plate} & 0.6 & - & - &-& 0.8 & - & - &-\\
            & Ours w/o symbol & 12.6 & 22.5 & 0.990 & 2.710& 6.9 & 11.7 & 0.972 & 2.917 \\
            & Ours w/o causal & 0.0 & - & - & 3.467 & 0.0 & - & - & 3.183\\
            & Ours w/ RL & 1.9 & 5.3 & \textbf{1.000} & 3.484& 1.4 & 4.9 & \textbf{1.000} & 3.370 \\
            & \cellcolor{mygray}\textbf{Ours} & \cellcolor{mygray}\textbf{61.8} & \cellcolor{mygray}\textbf{66.9} & \cellcolor{mygray}0.960 & \cellcolor{mygray}\textbf{0.307} & \cellcolor{mygray}\textbf{29.1} & \cellcolor{mygray}\textbf{43.9} & \cellcolor{mygray}0.954 & \cellcolor{mygray}\textbf{0.424}\\
            \toprule
            &\multicolumn{1}{c}{} &
            \multicolumn{4}{c}{Dataset level-1} &
            \multicolumn{4}{c}{Dataset level-2} \\
            
            \cmidrule{2-10}
            \multirow{5.5}{*}{\rotatebox[origin=c]{90}{\textbf{Unseen Task}}}
            & Chance & 0.4 & 2.1 & - & 3.550& 0.1 & 0.3 & - & 3.513  \\
            & PlaTe~\cite{sun2022plate} & 1.4 & - & - & - & 0.5 & - & - & - \\
            & Ours w/o symbol & 63.1 & 78.0 & 0.974 & 1.022& 40.0 & 51.9 & 0.980 & 1.407 \\
            & Ours w/o concept & 42.7 & 70.7& 0.971 & 1.485&-&-&-&- \\
            & Ours w/o causal & 0.0 & - & - & 3.536 & 0.0 & - & - & 3.525  \\
            & Ours w/ RL & 26.3 & 30.1 & \textbf{0.994} & 2.159& 2.8 & 7.0 & \textbf{1.000} & 3.417  \\
            & \cellcolor{mygray}\textbf{Ours} & \cellcolor{mygray}\textbf{98.7} & \cellcolor{mygray}\textbf{99.3} & \cellcolor{mygray}0.985 & \cellcolor{mygray}\textbf{0.015} & \cellcolor{mygray}\textbf{98.2} & \cellcolor{mygray}\textbf{99.4} & \cellcolor{mygray}0.991 & \cellcolor{mygray}\textbf{0.019} \\

            \toprule
           \multirow{9}{*}{\rotatebox[origin=c]{90}{\textbf{Real-world Data}}}
            &\multicolumn{1}{c}{} &
            \multicolumn{4}{c}{Dataset level-1} &
            \multicolumn{4}{c}{Dataset level-2} \\
            \cmidrule{2-10} 
            &Chance & 2.0 & 5.0 & -  & 3.261 & 1.0 & 2.0 & - &  3.370 \\
            &PlaTe~\cite{sun2022plate} & 12.0 & - & - & - & 5.0 & - & - & -  \\
            &\cellcolor{mygray}\textbf{Ours}  & \cellcolor{mygray}\textbf{52.0} & \cellcolor{mygray}\textbf{71.0} & \cellcolor{mygray}\textbf{0.980} & \cellcolor{mygray}\textbf{1.341} & \cellcolor{mygray}\textbf{36.0} & \cellcolor{mygray}\textbf{47.0} &\cellcolor{mygray}\textbf{0.987} & \cellcolor{mygray}\textbf{1.765}   \\
            \cmidrule{2-10} 
            &\multicolumn{1}{c}{} &
            \multicolumn{4}{c}{Dataset level-3} &
            \multicolumn{4}{c}{Dataset level-4} \\
            
            \cmidrule{2-10} 
            &Chance & 0.0 & 1.0 & - & 3.498  & 0.0 & 0.0 & - & 3.552  \\
            &PlaTe~\cite{sun2022plate} & 1.0 & - & - & -  & 1.0 & - & - & -   \\
            &\cellcolor{mygray}\textbf{Ours} & \cellcolor{mygray}\textbf{21.0}& \cellcolor{mygray}\textbf{27.0} &\cellcolor{mygray}\textbf{0.993}  & \cellcolor{mygray}\textbf{1.436}& \cellcolor{mygray}\textbf{11.0}& \cellcolor{mygray}\textbf{15.0} & \cellcolor{mygray}\textbf{1.000} &\cellcolor{mygray}\textbf{1.735} \\ 
            
            \bottomrule
        \end{tabular}%
        }
    \label{tab:gen_test}
\end{table}

We design three experiments to test the generalizabiliy of our model.

\paragraph{Unseen Objects} Through this experiment, we aim to investigate if our model can perform visual planning tasks on objects unseen during training. We test our model on the \textit{Unseen Object} testing dataset (see \cref{subsec:planning dataset} for details) and compare the results with several baselines to demonstrate the generalizability of our concept-based object representation module. We expect our concept learner to recognize the color, position, and size attributes of unseen object types during testing. If this is the case, the transition model could consequently apply transitions on these visual attributes for successful manipulation tasks. As shown in \cref{tab:gen_test}, our model is significantly more robust than PlaTe and RL-based methods against novel objects. 



\paragraph{Unseen Tasks} Moreover, we aim to verify that our model is flexible in processing atomic actions. We train our model on tasks with only limited types of action combinations, \ie, the \textit{Unseen Task} dataset. In this experiment, PlaTe only performs at the same level as a random guess, while our model performs as well as it does when being trained on the whole dataset (see \cref{tab:gen_test}), which demonstrates the generalizability of our method on unseen tasks.

\paragraph{Real-world Data} Finally, we assess our model's potential to generalize to real-world data. We collect a dataset of real images using Intel Realsense D455, which consists of four subsets, each representing a different difficulty level and comprising 100 tasks (equivalent to one-tenth of our validation sets' size). The real-world tasks are the same as the test tasks in dataset CCTP.
The quantitative results are demonstrated in \cref{tab:gen_test}. Since the real-world data and the \textit{CCTP} dataset have inherent discrepancies, our model, which was not finetuned with real-world data, exhibited a reduction in ASAcc and FSD. However, our model can successfully identify objects' color, position, and size within the real-world images, and outperform all the comparison models.
The qualitative results are shown at the bottom of \cref{fig:plan exp}. To simplify implementation, we focus on visually planning the target objects in the real-world images and ignore encoding and decoding the background.

\section{Conclusion}
\label{sec:conclusion}

In this paper, we propose a novel visual planning model based on concept-based disentangled representation learning, symbolic reasoning, and visual causal transition modeling.
In the future, we plan to extend our model to more complex planning tasks with diverse concepts and actions, and assist robots in real down-stream application tasks.

\textbf{Acknowledgement:} The authors would like to thank Mr. Zhitian Li and Dr. Meng Wang at BIGAI for their help with experiment setup. 

\small
\setstretch{0.96}
\bibliographystyle{IEEEtran}
\bibliography{reference}

\end{document}